\documentclass{article}

\usepackage[final]{corl_2019} 
\usepackage{graphicx}
\graphicspath{{Figures}}
\DeclareGraphicsExtensions{.pdf,.png}
\usepackage{amsthm}
\usepackage{multirow}
\newtheorem{theorem}{Theorem}
\newtheorem{proposition}{Proposition}
\usepackage{multirow}
\usepackage{bigstrut}

\newtheoremstyle{exampstyle}
{2pt} 
{2pt} 
{\itshape} 
{} 
{\bfseries} 
{.} 
{.5em} 
{\thmname{#1}\thmnumber{#2}\thmnote{(#3)}} 
\theoremstyle{exampstyle}
\newtheorem{hypothesis}{H}

\usepackage{amsmath}
\usepackage{amssymb}
\usepackage{bbm}

\usepackage{makecell}
\setcellgapes{1pt}
\usepackage[small]{caption}

\newcommand{\shrink}{\vspace{-10pt}}
\newcommand{\shrinktop}{\vspace{-13pt}}
\newcommand{\shrinkbottom}{\vspace{-7pt}}

\newcommand{\eref}[1]{Eqn.~(\ref{#1})}  
\newcommand{\sref}[1]{Sec.~\ref{#1}}    
\newcommand{\figref}[1]{Fig.~\ref{#1}}  
\newcommand{\tabref}[1]{Table~\ref{#1}} 
\newcommand{\prg}[1]{\noindent\textbf{#1. }} 

\newcommand{\mx}[1]{\textcolor{blue}{#1}}

\usepackage[colorinlistoftodos]{todonotes}

\usepackage{titlesec}
\titlespacing*{\subsection}{0pt}{0.1\baselineskip}{0.05\baselineskip}
\titlespacing*{\section}{0pt}{0.2\baselineskip}{0.1\baselineskip}

\definecolor{teal}{rgb}{0.26,0.52,0.56}
\definecolor{orange}{rgb}{0.89,0.55,0.06}

\title{Learning from My Partner's Actions: \\ Roles in Decentralized Robot Teams}

\author{
  Dylan P. Losey*\\
  Stanford University \\
  \texttt{dlosey@stanford.edu} \\
   \And
   Mengxi Li\thanks{Dylan P. Losey and Mengxi Li contributed equally to this work.} \\
   Stanford University \\
   \texttt{mengxili@stanford.edu} \\
   \AND
   Jeannette Bohg \\
   Stanford University \\
   \texttt{bohg@stanford.edu} \\
   \And
   Dorsa Sadigh \\
   Stanford University \\
   \texttt{dorsa@cs.stanford.edu} \\
}

\begin{document}
\maketitle



\vspace{-2em}
\begin{abstract}
    When teams of robots collaborate to complete a task, communication is often necessary. Like humans, robot teammates should \textit{implicitly} communicate through their \textit{actions}: but interpreting our partner's actions is typically difficult, since a given action may have many different underlying reasons. Here we propose an alternate approach: instead of not being able to infer whether an action is due to exploitation, information giving, or information gathering, we define separate \textit{roles} for each agent. Because each role defines a distinct reason for acting (e.g., only exploit), teammates can now correctly interpret the meaning behind their partner's actions. Our results suggest that leveraging and alternating roles leads to performance comparable to teams that explicitly exchange messages.
\end{abstract}



\section{Introduction}

\begin{figure}[t]
\shrink
	\begin{center}
		\includegraphics[width=1.0\columnwidth]{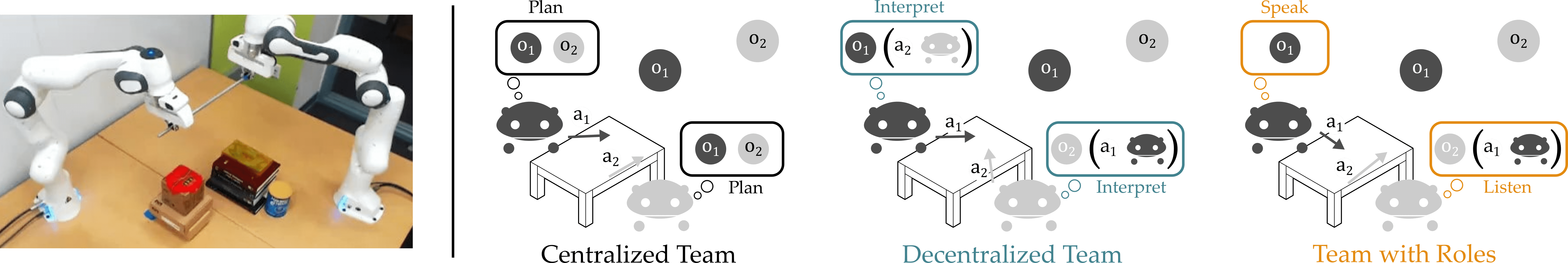}
		\caption{(Left) Our problem setting, where two robots must work together to complete a task. Each robot observes different parts of the true system state. (Right) Unlike \textbf{centralized teams}, \textbf{\textcolor{teal}{decentralized teammates}} should implicitly communicate through actions: here each agent sees one obstacle, and tries to infer where the other obstacle is based on their partner's actions. Interpreting our partner's actions is hard: there are many reasons why an agent might choose an action. We introduce speaker and listener  \textbf{\color{orange}{roles}} so that each agent has a distinct reason for acting, enabling the robots to learn from and implicitly communicate through their actions.}
		\label{fig:front}
	\end{center}
\vspace{-2em}
\end{figure}

When teams of robots are deployed in our homes, warehouses, and roads, often these robots must collaborate to accomplish their task. Imagine two robots working together to move a heavy table across a room (see \figref{fig:front}). Due to occlusions, each agent can only see some of the obstacles within this room. Thus, the robots need to \textit{communicate} to inform their partner about the obstacles they see. One option is for the robots to explicitly communicate by directly sending and receiving messages: i.e., we could tell our teammate where the obstacles are. But humans utilize more than just explicit communication---we also implicitly communicate through our actions. For example, if our partner guides the table away from our intended trajectory, we might infer that we were moving towards an obstacle, and that there is a better path to follow. Collaborative robot teams---like humans---should also leverage the information contained within their partner's actions to learn about the world.

Unfortunately, interpreting the meaning behind an action is hard. Robots can take actions for \textit{many different reasons}: to exploit what they know, actively give information to their partner, or actively gather information about the environment. So when we observe our partner applying some force to the table, what (if anything) should we learn from that action? And how do we select actions that our partner can also interpret? In this paper, we show that assigning roles alleviates these challenges: \vspace{-0.5em}
\begin{center}
\emph{Teams of robots can correctly interpret and learn from each other's actions when the team is} separated \emph{into} roles\emph{, and each role provides the robots with a distinct reason for acting.}
\end{center}\vspace{-0.5em}
Returning to our example, imagine that our partner's role is to exploit their current information: they move the table towards the goal while avoiding the obstacles that they can observe. If we know this role, we can now interpret their actions: when our partner applies an unexpected force, it must be because of some obstacle that we did not see. Hence, assigning roles enables us to \textit{learn} from our partner's actions and update our estimate of the system state \textit{naturally}, without requiring additional, explicit communication. In this paper, we make the following contributions:

\textbf{Roles in Two-Player Teams.} We focus on decentralized two-player teams where each agent sees part of the current state, and together the agents observe the entire state. We show that---without roles---the agent policies are interdependent, and interpreting the actions of our partner leads to infinite recursion (what do you think I think you think, etc.). To remove this interdependence and reach interpretable actions, we introduce two classes of policies: a \textit{speaker} role, where agents exploit what they know, and a \textit{listener} role, where agents learn by modeling their partner as a speaker.

\textbf{Mimicking Explicit Communication.} We explore how roles can be used to make our decentralized team behave like a \textit{centralized} team that communicates explicitly. We find that decentralized teammates which alternate between roles can match the centralized team, but if the agents always maintain the same roles, the team may become unstable. We also reveal that speakers trade-off between stochasticity and communication: to improve overall team performance, speakers should choose more deterministic actions than the centralized policy to clearly convey their observations.

\textbf{Comparing Implicit to Explicit.} We implement roles both in a simulated game and a two-robot manipulation task. Our simulations compare implicitly communicating through actions to explicitly communicating by sending messages, and demonstrate that---when robots leverage roles---implicit communication is almost as effective as explicit communication. In robot experiments, teams that alternated roles successfully communicate their observations and collaborate to avoid obstacles.

Overall, this work is a step towards learning from our partner's actions in decentralized robot teams.

\section{Related Work}

\textbf{Multi-Agent Teams.} 
Teams of robots have performed tasks such as localization \cite{prorok2012low}, navigation \cite{ayanian2010decentralized},
and manipulation \cite{song2002potential, sreenath2013dynamics}. While many works rely on centralized coordination \cite{bowman2016coordination}, decentralized multi-agent teams are receiving increasing attention  \mx{\cite{culbertson2018decentralized, ayanian2010decentralized, nigam2018sufficient, lessard2015optimal}}. Within \textit{decentralized control}, the problem is reformulated as a decentralized partially-observable Markov decision process (Dec-POMDP) \cite{amato2013decentralized, bernstein2002complexity} or coordinator-POMDP \cite{nayyar2013decentralized}. 
In practice, solving these optimization problems is frequently intractable: related works instead rely on high-level abstractions \cite{amato2015planning}, sparse interactions between agents \cite{melo2011decentralized}, specific problem instances \cite{swigart2011optimal}, or control approximations \cite{culbertson2018decentralized}. Alternatively, with multi-agent \textit{reinforcement learning} (MARL), agents can learn decentralized policies from trial-and-error \cite{bu2008comprehensive}. Today's MARL approaches often leverage actor-critic methods to scale to high dimension state-action spaces \cite{lowe2017multi, gupta2017cooperative, foerster2018counterfactual}; however, these methods require offline, centralized training, and must deal with the credit assignment problem (i.e., who is contributing to the team's success?).

\textbf{Communication.} Across both control and learning approaches, communication between agents is key to effective collaboration. The protocol that the agents use to communicate can be either based on pre-defined triggers \cite{seyboth2013event, johnson2016role} or learned from training data \cite{sukhbaatar2016learning, foerster2016learning, paulos2019decentralization}. But the method that the agents use to communicate is generally \textit{explicit}: the agents have a separate channel with which they directly broadcast and receive messages from their teammates \cite{yan2013survey}. At the other end of the spectrum, \cite{wang2016kinematic, pagello1999cooperative, zhang2017role} study cooperative robots that do \textit{not} attempt to exchange information. Unlike these recent works that either have access to explicit communication or omit communication entirely, we will focus on leveraging \textit{implicit} communication through actions.

\textbf{Roles in Human Teams.} Our research is inspired by human-human teams, where different roles naturally emerge in collaborative tasks \cite{nalepka2017herd, hawkins2018emergence}. For example, when two humans are working together to physically manipulate an object, the agents can identify and assume complementary roles based on their partner's force feedback alone \cite{reed2008physical}. Roles have also been applied to human-robot interaction: here the robot dynamically adjusts its level of autonomy (i.e., becomes a leader or follower) in response to the force feedback from the human partner \cite{mortl2012role, losey2018review, hadfield2016cooperative}. We will \textit{extend} these ideas to robot-robot teams, where we believe that roles can similarly improve collaboration.

\section{Interpreting Actions via Roles}
\label{sec:roles_intro}

Consider a decentralized, two-agent team where the agents share a common objective. Each agent observes part of the system state. To make good decisions, both agents need an accurate estimate of the entire state; for instance, we need to know whether there is an obstacle behind us, if there is free space to our right, etc. While we know the aspects of the state that we directly observe, how can we estimate the parts of the system state that our partner sees? One natural solution is to learn about the environment based on the \textit{implicit} communication contained within our partner's \textit{actions.} In this section, we show that correctly interpreting the meaning behind our partner's actions is challenging when both agents try to learn from their partner and exploit what they observe at the same time. Returning to the table moving example: when our partner applies a force, is this because of what they have learned from our own actions, because of an obstacle behind us, or some combination of both? To correctly infer the unobserved state, each agent must reason over their partner's behavior, and this behavior may in turn depend on the first agent's actions. Accordingly, we here introduce \textit{speaker} and \textit{listener} roles to remove this interdependency: the speaker implicitly communicates relevant parts of the state that they observe to the listener, who learns a more accurate state estimate.

\textbf{Two-Agent Team.} We formulate our two-player team as a decentralized Markov decision process (Dec-MDP) \cite{amato2013decentralized}. Let the state space be $\mathcal{S} = \mathcal{S}_1 \times \mathcal{S}_2$ and let the action space be $\mathcal{A} = \mathcal{A}_1 \times \mathcal{A}_2$, where $\mathcal{S}_i$ and $\mathcal{A}_i$ are the state and action space for agent $i$. The team has dynamics $T : \mathcal{S} \times \mathcal{A} \times \mathcal{S} \rightarrow [0,1]$ and receives reward $R : \mathcal{S} \times \mathcal{A} \rightarrow \mathbb{R}$. At the current timestep $t$, agent $i$ observes its own state component $s_i^t$, and collectively the team observes $s^t = (s_1^t, s_2^t)$. We therefore have that the state is \textit{jointly} fully observable: $s^t$ is known given the current observations of both agents, $s_1^t$ and $s_2^t$. When making decisions, agent $i$ has access to its history of observations, $s_i^{0:t} = (s_i^0, s_i^1, \ldots, s_i^t)$, as well as the history of actions taken by both agents, $a^{0:t-1}$. For simplicity, we assume that each agent observes its partner's current action selection (our results still hold if they observe the previous action). Hence, the agents have policies of the form $\pi_1(a_1^t \mid s_1^{0:t},a^{0:t-1},a_2^t)$ and $\pi_2(a_2^t\mid s_2^{0:t},a^{0:t-1}, a_1^t)$.

\textbf{Mimicking Centralized Teams.} One approach to control decentralized teams is solving this Dec-MDP; however, the problem is NEXP-complete \cite{bernstein2002complexity}, and often intractable for continuous state and action spaces. We study a different approach: we find policies for both decentralized agents to collectively \textit{mimic} the behavior of a \textit{centralized} team \cite{dobbe2017fully, paulos2019decentralization}. Imagine that---when moving a table---both teammates know exactly what their partner sees; when we explicitly communicate our observations, we can solve the problem together, and collaborate perfectly to carry the table. We will treat this centralized policy that uses explicit communication as the \textit{gold standard} for our decentralized team, while recognizing that the decentralized team only has access to implicit communication, and may not be able to completely match the centralized team. Importantly, the optimal centralized policy is the solution to an MDP, and can be tractable computed offline (P-complete) \cite{bernstein2002complexity}.

\textbf{Interdependent Policies.} Let the centralized policies be $\pi_1^*(a_1^t~|~s_1^t,s_2^t)$ and $\pi_2^*(a_2^t~|~s_1^t,s_2^t)$. When both decentralized agents choose their actions to best mimic this centralized behavior, we reach:
\begin{equation}
\label{eq:P1}
\begin{gathered}
    \pi_1(a_1^t~|~s_1^{0:t},a^{0:t-1}, a_2^t) \propto \sum_{s_2^{0:t}} \pi_1^*(a_1^t~|~s_1^t,s_2^t) \cdot \pi_2(a_2^t~|~s_2^{0:t},a^{0:t-1},a_1^t) \cdot P(s_2^{0:t}~|~s_1^{0:t},a^{0:t-1}) \\
    \pi_2(a_2^t~|~s_2^{0:t},a^{0:t-1},a_1^t) \propto \sum_{s_1^{0:t}} \pi_2^*(a_2^t~|~s_1^t,s_2^t) \cdot \pi_1(a_1^t~|~s_1^{0:t},a^{0:t-1},a_2^t) \cdot P(s_1^{0:t}~|~s_2^{0:t},a^{0:t-1})
\end{gathered}
\end{equation}

See the appendix for our complete derivation. We cannot solve Eqn.~(\ref{eq:P1}) as is because the policies are interdependent: $\pi_1$ and $\pi_2$ both appear in each other's policy, so that solving for $\pi_1$ requires solving $\pi_2$ for all possible $s_2^{0:t}$, which requires an inner loop that solves for $\pi_1$ over all possible $s_1^{0:t}$, and so on. Going back to our table example: imagine that our partner observes whether there is an obstacle behind us, and we want to infer the likelihood of this obstacle from their actions. This is easy when our partner's actions are only based on this obstacle---but if our partner's behavior is also a response to our own actions, how to we know which aspects of our partner's behavior to learn from? To break this interdependence and recover interpretable actions, we separate our team into two roles: an agent that exploits what it observes (the \textit{speaker}), and an agent that learns from its partner's (the \textit{listener}).

\textbf{Speaker.} A speaker is an agent that makes decisions by purely exploiting their observations $s_i^{0:t}$. Let Agent $1$ be the speaker; the speaker policy that best matches $\pi_1^*$ is:
\begin{equation} \label{eq:P2}
    \pi_1(a_1^t~|~s_1^{0:t}) = \sum_{s_2} \pi_1^*(a_1^t ~|~ s_1^t,s_2^t) \cdot P(s_2^t~|~s_1^{0:t})
\end{equation}
More generally, any policy of the form $\pi_i(a_i^t~|~s_i^{0:t})$ is a speaker. Because speakers react to their observations, these actions convey information about the parts of the state they see to their teammates.

\textbf{Listener.} A listener makes decisions based on its own observations while also learning an improved state estimate from its partner's behavior. If Agent $2$ is the listener, the policy that matches $\pi_2^*$ is:
\begin{equation} \label{eq:P3}
    \pi_2(a_2^t~|~s_2^{0:t},a^{0:t-1},a_1^t) \propto \sum_{s_1^{0:t}} \pi_2^*(a_2^t~|~s_1^t,s_2^t) \cdot \pi_1(a_1^t~|~s_1^{0:t}) \cdot P(s_1^{0:t}~|~s_2^{0:t},a^{0:t-1})
\end{equation}
Compared to (\ref{eq:P1}), now the listener models its partner as a speaker, and solving for $\pi_1$ (the speaker policy) does not depend on $\pi_2$ (the listener policy). More generally, we consider any policy of the form $\pi_i(a_i^t~|~s_i^{0:t},a^{0:t-1}, a_j^t)$ as a listener if it interprets its teammate's actions with $\pi_j(a_j^t~|~s_j^{0:t})$.

\section{Leveraging Roles Effectively}
\label{sec:roles_use}

Now that we have defined roles, let us return to our table carrying example. Imagine that we are the speaker and our partner is the listener: what is the best speaker policy for us to follow? Should we always remain a speaker, or do we need to switch speaker and listener roles with our teammate? And if our decentralized team uses roles, when can we fully match the behavior of a centralized team? In this section we explore these questions, and analyze how roles operate within simplified settings.

\subsection{When Can We Use Roles to Match a Centralized Team?}

Roles enable implicit communication through actions. This implicit communication is typically less informative than explicitly sharing observations; but when an agent's actions can completely convey their observed state, robots can leverage roles to fully match the behavior of a centralized team:
\begin{theorem} \label{theorem1}
    For continuous systems, if there exist surjective functions ${g_1 : \mathcal{A}_1 \rightarrow S_1}$ and ${g_2 : \mathcal{A}_2 \rightarrow S_2}$, a decentralized team using speaker and listener roles can match a centralized team. 
\end{theorem} \vspace{-1.0em}
\begin{proof}
The decentralized team matches the centralized team's performance by communicating with actions while rapidly alternating between speaker and listener roles. Define $a_1^*$ as the optimal centralized action for agent $1$, and let $\bar{a}_1$ be a naive action that completely conveys what agent $1$ observes: $g_1(\bar{a}_1) = s_1$. We choose the speaker action to be $\bar{a}_1$ and the listener action to be $a_2^* + (a_2^* - \bar{a}_2)$, where the listener can compute $a_2^*$ because it observes $s_2$ and infers $s_1$ from $g_1(\bar{a}_1)$. The agents change roles: agent $1$ is the speaker for time $[t, t + dt)$ and agent $2$ is the speaker for time $[t+dt, t+2dt)$. Taking the limit as $dt \rightarrow 0$ (i.e., as the roles change infinitely fast), the team's action $a$ becomes:
\begin{equation*}
\label{eq:L1}
    a^t = \lim_{dt\rightarrow 0} \frac{a^t + a^{t+dt}}{2} = \lim_{dt \rightarrow 0}\frac{1}{2}
    \bigg(\begin{bmatrix} \bar{a}_1^t \\ a_2^{*t} + (a_2^{*t} - \bar{a_2}^t)\end{bmatrix} + \begin{bmatrix} a_1^{*t+dt} + (a_1^{*t+dt} - \bar{a}_1^{t+dt}) \\ \bar{a}_2^{t+dt}\end{bmatrix}\bigg) = a^{*t}
\end{equation*}
and so the decentralized team's action converges to the optimal action of the centralized team.
\end{proof} \vspace{-1.0em}
Intuitively, consider the table example with two dynamic obstacles, one of which is observed by each teammate. When we are the speaker, we apply a force with a direction and magnitude that conveys our obstacle's position to the partner; likewise, when our partner is the speaker, their action informs us where their obstacle is. By quickly switching speaker and listener roles we can both understand the state, and match the behavior of a team that explicitly tells one another about the obstacles. Theorem \ref{theorem1} can also be extended to a team of $N$ agents (see appendix for full proof).

\subsection{Analyzing Roles in Linear Feedback Systems}

To better understand how roles affect decentralized teams, we specifically focus on teams controlled using linear feedback. Here the centralized policy is $a^* = -K^*s$, where $K^*$ is the desired control gain; for instance, this policy could be the solution to a linear-quadratic regulator (LQR). Assuming our decentralized team is similarly controlled with $a = -Ks$, we first determine whether alternating speaker and listener roles is necessary to ensure system stability. We next consider situations where the centralized policy includes stochastic behavior, and we identify how speakers should optimally trade-off between mimicking this desired noise and effectively communicating with the listener. 

\textbf{Do We Need to Change Roles?} Imagine that we are the speaker within the table-carrying example. In the best case, our actions completely convey our observations to our partner. But if we are never a listener, we never know what our partner observes; and, if the team's behavior depends on our own actions, this can lead to situations where we are unable to collaboratively accomplish the task:
\begin{proposition} \label{prop1}
    If the teammates never change speaker and listener roles, and the team is attempting to mimic a centralized controller $a^* = -K^*s$, there exist controllable system dynamics for which the decentralized team $a = -Ks$ becomes unstable for any choice of $K$.
\end{proposition} \vspace{-1.0em}
\begin{proof}
    We prove this by example. Let agent $1$ always be the speaker and let agent $2$ be the listener. Consider the following controllable system with linear dynamics $\dot{s} = As + Ba$, where $a = -Ks$:
    \begin{equation} \label{eq:L2}
        \begin{bmatrix} \dot{s}_1 \\ \dot{s}_2 \end{bmatrix} = \begin{bmatrix} 1 & 1 \\ 0 & 1 \end{bmatrix}\begin{bmatrix} s_1 \\ s_2 \end{bmatrix} - \begin{bmatrix} 0 & 0 \\ 1 & 0 \end{bmatrix}\begin{bmatrix} K_{11} & 0 \\ K_{21} & K_{22} \end{bmatrix}\begin{bmatrix} s_1 \\ s_2 \end{bmatrix}
    \end{equation}
    Note that $K_{12} = 0$; this is because the speaker makes decisions based purely on their own state, $s_1$. The listener is able to perfectly infer $s_1$ since it observes the speaker's action $a_1$, and $s_1 = -K_{11}^{-1}a_1$. For this team to have stable dynamics the eigenvalues of $A - BK$ must have a strictly negative real part. But here the eigenvalues are: $1 \pm \sqrt{-K_{11}}$. Hence, no matter what speaker and listener gains $K_{11}$, $K_{21}$, and $K_{22}$ we choose, the decentralized team becomes unstable.
\end{proof} \vspace{-0.5em}

\textbf{How Should We Speak and Listen with Noise?} Returning to our table carrying example, we now know that we should alternate speaker and listener roles even when our actions completely convey our observations. But what if we cannot fully observe the actions of our partner (e.g. due to sensor noise)? 
As long as this estimate is unbiased, we can still match the expected behavior of the centralized team by treating these 
noisy actions as accurate measurements: 
\begin{proposition} \label{prop2}
    If the teammates incorrectly sense their partner's action $a_i$ as $a_i + n_i$, where $n_i$ is unbiased noise, the decentralized team can match the centralized controller action $a^* = -K^*s$ in expectation by speaking and listening as if $a_i + n_i$ were the teammate's true action.
\end{proposition} \vspace{-1.0em}
\begin{proof}
    Let agent $1$ be the speaker, let agent $2$ be the listener, and let the decentralized controller be $a=-Ks$, where $K$ is shown in \eref{eq:L2}. The speaker takes action $a_1 = -K_{11}s_1$, but the listener measures $a_1 + n_1$, and infers the speaker's state as $\hat{s}_1 = s_1 -K_{11}^{-1}n_1$. Accordingly, when the listener acts based on $\hat{s}_1$, their action has an additional term $K_{21}K_{11}^{-1}n_1$. Next the agents switch roles, and the first agent (i.e., the listener) acts with an additional term $K_{12}K_{22}^{-1}n_2$. When the agents rapidly alternate between speaker and listener roles, the team's behavior includes these terms:
    \begin{equation} \label{eq:L3}
        a = -K^*s + \begin{bmatrix} K_{12}^*K_{22}^{*,-1}n_2 \\ K_{21}^*K_{11}^{*,-1}n_1 \end{bmatrix}, \quad K^* = \begin{bmatrix} K_{11}^* & K_{12}^* \\ K_{21}^* & K_{22}^* \end{bmatrix}, \quad \mathbb{E}_{n}[a] = -K^*s = a^*
    \end{equation}
    This matches the centralized team's action in expectation if $n_1$ and $n_2$ are unbiased.
\end{proof} \vspace{-1.0em}

Noisy action measurements are undesirable but often unavoidable. Conversely, there are cases where the centralized policy itself \textit{recommends} stochastic actions: i.e., when an agent sees an obstacle in front of it, it should go around on the right-side $30\%$ of the time, and on the left otherwise. The speaker and listener can \textit{choose} whether or not to incorporate this stochasticity. Intuitively, we might think that both the speaker and listener should match the optimal centralized policy; but, when the speaker's actions become more stochastic, it is harder for the listener to correctly interpret what the speaker has observed. Imagine we are the listener: the more random the speaker's action is,  the less information we have about what the speaker sees, which makes it more challenging for us to learn the system state. This leads to a trade-off between speaker noise and overall team performance:
\begin{proposition} \label{prop3}
    If the centralized policy is stochastic, so that $a^* = -K^*s + n$, and $n$ is sampled from a Gaussian distribution $n \sim \mathcal{N}(0, diag(w_1^2, w_2^2))$, the speaker policy that minimizes the Kullback-Leibler (KL) divergence between the decentralized and centralized policies has a variance less than or equal to the corresponding variance of the centralized policy.
\end{proposition} \vspace{-1.0em}
\begin{proof}
    Let the first agent be the speaker and let the second agent be the listener. The decentralized team takes action $a = -Ks + m$, where $m \sim \mathcal{N}(0, diag(\sigma_1^2,\sigma_2^2))$. We solve for the variances $\sigma_1^2$ and $\sigma_2^2$ that minimize the KL divergence between the decentralized and centralized policies:
    \begin{equation} \label{eq:L4}
        \min_{\sigma_1^2,\sigma_2^2} ~\text{KL}\Big[\mathcal{N}\big(-Ks,diag(\sigma_1^2,\sigma_2^2)\big),~ \mathcal{N}\big(-K^*s,diag(w_1^2,w_2^2)\big)\Big]
    \end{equation}
    Selecting $K$ to best match $K^*$, and taking the expectation of \eref{eq:L4}, the optimal variances are:
    \begin{equation} \label{eq:L5}
        \sigma_1^2 = \frac{K_{11}^2w_1^2w_2^2}{K_{11}^2w_2^2 + K_{21}^2w_1^2} \leq w_1^2, \quad \sigma_2^2 = w_2^2
    \end{equation}
    See the appendix for our full derivation. Hence, while the listener should match the variance of the corresponding centralized policy, the speaker should intentionally choose a variance $\sigma_1^2 \leq w_1^2$. 
\end{proof} \vspace{-0.5em}

\textbf{Summary.} If two decentralized agents are collaborating with roles, they can match the behavior of a centralized team when their actions completely convey their observations. But even in these cases, alternating between speaker and listener roles is necessary; the team may become unstable if their roles never change. When choosing how best to speak and listen, greedily mimicking the centralized policy is effective, and robust to noisy, unbiased measurements of the partner's actions. In situations where the centralized policy is stochastic, however, a trade-off emerges: speakers should select more deterministic actions in order to better convey their observations to the listener.

\section{Simulations and Experiments}

\begin{figure}[t]
    \shrink
	\begin{center}
		\includegraphics[width=0.8\columnwidth, height=100pt]{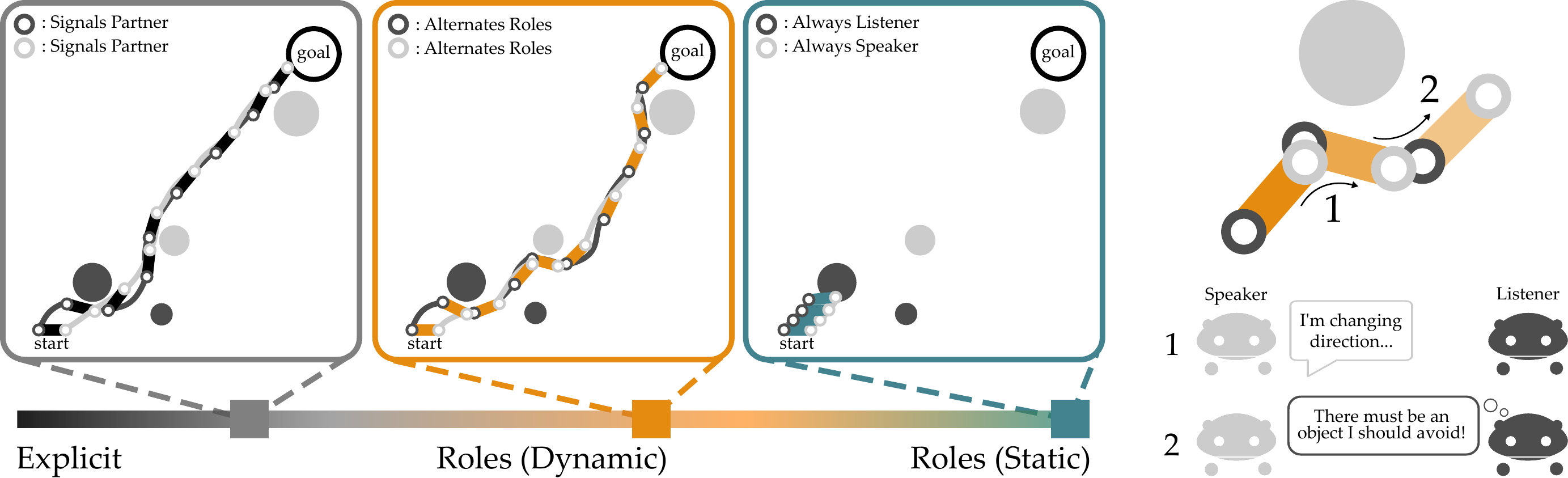}

		\caption{Simulated task. (Left) Two agents collaborate to carry a rigid object while avoiding obstacles. Each agent can only see the obstacles matching their color. In this example, the \textit{explicit} and \textit{dynamic roles} teams are able to negotiate the obstacles to reach the goal, while the \textit{static roles} team fails. (Right) Example of implicit communication via actions: here the speaker sees an obstacle, and abruptly moves to the right (1). The listener updates its state estimate based on this unexpected motion, and also moves to avoid the unseen obstacle (2).}

		\label{fig:game_fig}
	\end{center}
\vspace{-2.5em}
\end{figure}

Roles are sufficient when actions can completely convey the state an agent observes. But what about more general situations, where the observation space has a higher dimension than the action space? And how do our proposed roles function on actual robot teams? Here we explore these challenging cases in a simulated game (see \figref{fig:game_fig}) and a two-robot manipulation task (see \figref{fig:panda}). We compare fixed and dynamic role allocations to different amounts of explicit communication. Overall, we find that there is a \textit{spectrum} across explicit and implicit communication, and that implicit communication via roles approaches the performance of equivalent explicit communication.

\subsection{Simulated Table-Carrying Game with Implicit and Explicit Communication}

We simulated a continuous state-action task in which two point robots carried a table across a plane (see Fig.~\ref{fig:game_fig}). Each robot observed half of the obstacles within this plane, and so together the team needed to communicate to obtain an accurate state estimate. We compared using \textit{explicit} and \textit{implicit} communication. During explicit communication, agents sent and received messages that contained the exact location and geometry of the closest obstacle. By contrast, during implicit communication the agents leveraged roles to learn from their partner's actions. An example is shown in Fig.~\ref{fig:game_fig}: here the speaker sees an obstacle---that the listener cannot observe---and changes its motion to implicitly indicate to the listener that there is an obstacle directly ahead.

\textbf{Independent Variables.} We varied the (a) communication strategy and (b) task complexity. There were three levels of communication strategy: \textit{explicit}, \textit{dynamic roles}, and \textit{static roles}. Within \textit{explicit} and \textit{dynamic roles}, we also varied the number of timesteps $T$ between communication; i.e., the \textit{explicit} teams could only send messages about the closest obstacle every $T$ timesteps, or, analogously, the speaker and listener roles alternated after $T$ timesteps. In order to adjust the task complexity, we increased $n$, the total number of obstacles in the environment.

\textbf{Dependent Measures.} Within each simulation, two agents held a rigid table, and tried to carry that table to the goal without colliding with an obstacle. We measured the \textit{success rate} ($\lambda$) of reaching the goal over $1000$ randomly generated environments. As a baseline, we also tested fully centralized teams: these centralized teams reached the goal in all environments ($\lambda = 1$).

\textbf{Hypotheses:} 
\begin{hypothesis}
    Dynamically alternating roles leads to better performance than fixed role allocations, even in cases where actions can completely convey the location of the closest obstacle. 
\end{hypothesis}
\begin{hypothesis}
    In environments where actions cannot completely convey the closest obstacle, dynamic role allocations perform almost as well as explicitly communicating the closest obstacle.
\end{hypothesis}
\begin{hypothesis}
    Implicit communication via roles is robust to noisy action observations.
\end{hypothesis}

We provide additional details about this simulated game in the Appendix (Sec.~\ref{sec:appendix}).

\subsection{Rapidly Changing Roles Outperforms Static Roles}
\label{sec:exp_1}
\begin{table*}[t]
\footnotesize
\shrink
	\caption{Success rate $\lambda$ when both robots knew the geometry of the obstacles \textit{a priori}, but not their locations. Each agent implicitly communicated the positions of the obstacles that they could see through their actions. Importantly, the agents could completely communicate the position of the closest obstacle with their current action: hence, the success rate of \textit{explicit} teams (not listed) is almost identical to that of \textit{dynamic} teams.} \vspace{-0.5em}
	\label{table}
	\begin{center}
		\begin{tabular}{{c}|{c}{c}{c}{c}{c}}
 & \multicolumn{3}{c}{\textbf{Roles (Dynamic)}} & \multicolumn{2}{c}{\textbf{Roles (Static)}} \\ 
Obstacles & \textbf{$T=1$} & \textbf{$T=4$} & \textbf{$T=16$} & Speaker-Listener & Speaker-Speaker \bigstrut \\ \hline
$n=2$ & $0.995$ & $0.914$ & $0.827$ & $0.872$ & $0.757$ \bigstrut[t] \\ $n=4$ & $0.982$ & $0.786$ & $0.656$ & $0.735$ & $0.545$ \\ $n=8$ & $0.924$ & $0.586$ & $0.434$ & $0.539$ & $0.305$
		\end{tabular}
		\label{table:table1}
	\end{center}
	\vspace{-2.7em}
\end{table*}

We first tested hypothesis \textbf{H1} in a setting where each agent knew the geometry of all the obstacles in the environment (i.e., all obstacles were circles with the same radius). Since the obstacle shape is known and fixed, agents could fully convey the closest obstacle's $(x,y)$ location through their $2$-DoF actions. Our results are shown in Table~\ref{table:table1}; we point out that \textit{dynamic roles} and \textit{explicit} are almost the same in this case, and so we focus on comparing \textit{dynamic roles} to \textit{static roles}. For each tested number of obstacles, the teams that rapidly alternated roles outperformed teams with fixed roles. Indeed, when there are only $n=2$ obstacles, the mapping from action space to observation space was surjective, and the \textit{dynamic roles} team converged towards $\lambda = 1$ as $T \rightarrow 0$ (Theorem.~\ref{theorem1}). But dynamically alternating roles was not always better: when the teammates changed roles too slowly ($T=16$), their performance was actually worse than maintaining a constant speaker and listener.

\subsection{Implicitly Communicating via Roles Competes with Explicit Communication}
\label{sec:exp_2}
\begin{figure}[b]
    \shrink
	\begin{center}
		\includegraphics[width=0.9\columnwidth]{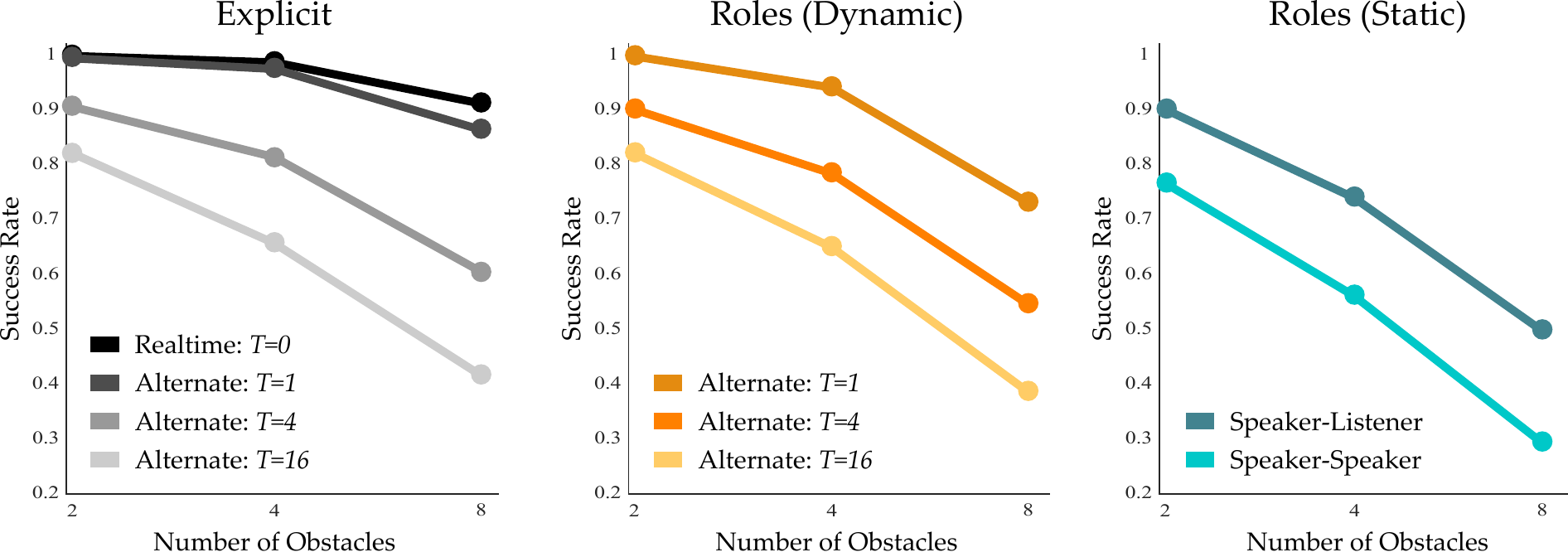}
		\caption{Success rate $\lambda$ when the robots did not know the obstacle geometry \textit{a priori}. Not only was the observation space higher dimensional than the action space, but agents could not even convey the closest obstacle's position and radius in a single action. There is a spectrum in performance across both communication type and frequency. For realtime explicit communication, the agents fully convey the closest obstacle at every timestep.}
		\label{fig:sim2}
	\end{center}
\shrink
\end{figure}

Next, we explored hypothesis \textbf{H2} in more complex settings where the agent's current action could not completely convey what they observed to their partner. Here the robots did not know the obstacle geometry \textit{a priori}; instead, each obstacle radius was randomly sampled from a uniform distribution. Teams with \textit{roles} had to try and implicitly communicate about both the obstacle location and shape---by contrast, \textit{explicit} teams could send messages to their partner that completely conveyed the closest obstacle. Our findings are visualized in Fig.~\ref{fig:sim2}. We see a spectrum in performance across \textit{explicit}, \textit{dynamic roles}, and \textit{static roles}. Although directly sending obstacle information is always better than learning from actions, implicitly communicating through \textit{dynamic roles} is almost as successful as \textit{explicitly} communicating with our partner at the same time interval $T$. This suggests that leveraging roles to learn from our partner's actions can be nearly as effective as direct communication, even in cases where the observations cannot be completely conveyed in a single action.

\subsection{Roles with Noisy Action Observations Match Noisy Explicit Communication}
\label{sec:exp_3}

So far we have conducted simulations in settings where the agents can perfectly measure the communication of their partner. Now we take a step back, and consider hypothesis \textbf{H3} when the communication channel \textit{itself} is noisy. This encompasses scenarios where the partners cannot perfectly measure each other's actions, or, analogously, when the explicit messages are corrupted. Our results with zero mean Gaussian noise are listed in Table~\ref{table:table2}. As expected, including noise decreased performance across both \textit{explicit} and \textit{dynamic roles}. But teams with \textit{dynamic roles} that rapidly alternated were still almost on par with \textit{explicit} teams that communicated in realtime, demonstrating that roles were as robust to noisy actions as explicit teams were to noisy messages. Similar to before, fixed speaker-listener teams were more successful than teams that slowly alternated roles.

\begin{table*}[t]
\footnotesize
\shrinktop
	\caption{Success rate $\lambda$ when the communication was noisy. We simulated zero-mean Gaussian noise, and chose the variance so that the coefficient of variation was $0.1$. Both explicit communication (noisy messages) and implicit communication (noisy action observations) had the same noise ratio. Dynamic roles performed almost on par with realtime explicit communication, where both agents exchanged messages at every timestep.}\vspace{-0.5em}
	\label{table}
	\begin{center}
		\begin{tabular}{{c}|{c}{c}{c}{c}{c}{c}}
& \textbf{Explicit} & \multicolumn{3}{c}{\textbf{Roles (Dynamic)}} & \multicolumn{2}{c}{\textbf{Roles (Static)}} \\ 
Obstacles & $T=0$ & $T=1$ & $T=4$ & $T=16$ & Speaker-Listener & Speaker-Speaker \bigstrut \\ \hline
$n=2$ & $0.925$ & $0.884$ & $0.818$ & $0.787$ & $0.829$ & $0.757$ \bigstrut[t] \\ $n=4$ & $0.833$ & $0.747$ & $0.669$ & $0.606$ & $0.655$ & $0.545$ \\ $n=8$ & $0.553$ & $0.512$ & $0.416$ & $0.357$ & $0.398$ & $0.305$
		\end{tabular}
		\label{table:table2}
	\end{center}
	\vspace{-2.5em}
\end{table*}

\subsection{Manipulation Experiments with Two Robot Arms}

\begin{figure}[b]
    \shrink
    \vspace{-0.5em}
	\begin{center}
		\includegraphics[width=0.8\columnwidth]{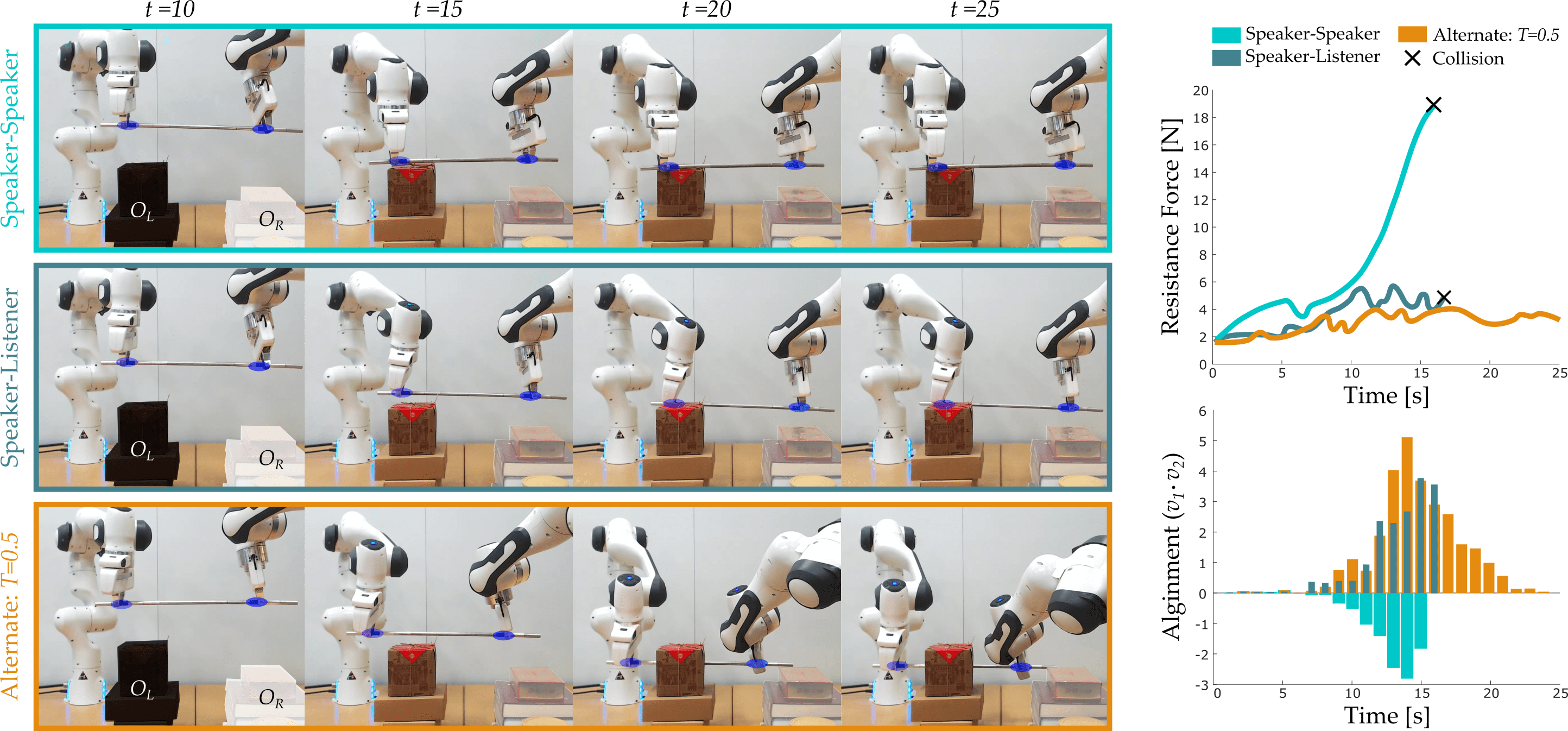}

		\caption{Two decentralized robot arms tasked with placing a rod on the table while implicitly communicating via roles: the left robot sees obstacle $O_L$ and the right robot sees $O_R$. (Left) The behavior of \textit{static role} and \textit{dynamic role} teams after $t$ s. Both static role teams collided with obstacle $O_L$ because they failed to mutually communicate their observations. The team that alternated roles successfully reasoned over each other's actions, conveyed the obstacle positions, and aligned their actions. (Right) We plot the norm of the force the two agents exerted on one another, as well as the alignment between the two agent's end-effector movements.}

		\label{fig:panda}
	\end{center}
    \vspace{-1em}
\end{figure}

We implemented our \textit{dynamic role} and \textit{static role} strategies on two decentralized robot arms (Panda, Franka Emika). Our experimental setup is shown in Fig.~\ref{fig:panda}: the robots were controlled using separate computers, and were tasked with placing a rod on the table top without any explicit communication. In order to complete this task, the robots had to negotiate two obstacles---but, like in our simulations, each robot could only see one of these obstacles. Because of their differing knowledge about the system state, the robots originally had opposite plans about how to move the rod to the table: one robot wanted to move the rod forwards (with respect to the camera), while the other robot planned to move the rod backwards. We expect intelligent robots to recognize that there is a \textit{reason} why their partner disagrees with them, and \textit{learn} from their partner's actions in realtime to update their estimate of the system state. We controlled the robots using static speaker-speaker and speaker-listener roles, as well as dynamic roles that alternated every $0.5$ s. During the experiments, only the \textit{dynamic roles} team inferred the obstacles that their partners observed, and aligned their actions to successfully place the rod on the table (see Fig.~\ref{fig:panda} and \href{http://ai.stanford.edu/blog/learning-from-partners/}{\color{blue}{ai.stanford.edu/blog/learning-from-partners/}}).
\section{Discussion}

\prg{Summary} Decentralized robot teams should learn about what their partner observes based on their partner's actions, but this is not possible when both agents attempt to simultaneously exploit their observations and communicate with their partner. We have therefore introduced separate speaker and listener roles: our analysis shows that teammates which dynamically alternate roles theoretically and experimentally approach the performance of teammates that can explicitly communicate.

\prg{Limitations and Future Work} This work is limited to Dec-MDPs, where the agents collectively observe the full system state. But we recognize that often there are parts of the state that neither agent can fully observe; accordingly, our future work will focus on extending roles to Dec-POMDPs.

\vspace{2em}
\textbf{Acknowledgements}

The authors would like to acknowledge NSF Award $\#1544199$. Toyota Research Institute (“TRI”)
provided funds to assist the authors with their research but this article solely reflects the opinions and conclusions of its authors and not TRI or any other Toyota entity.


\bibliography{citations}
\newpage \clearpage
\section{Appendix}
\label{sec:appendix}

\subsection{Derivations for \eref{eq:P1}}
\begin{equation}
\begin{split}
    \pi_1(a_1^t~|~s_1^{0:t},a^{0:t-1}, a_2^t) &=  \sum_{s_2^{0:t}} P(a_1^t, s_2^{0:t}~|~s_1^{0:t},a^{0:t-1}, a_2^t) \\
    & = \frac{\sum_{s_2^{0:t}} P(a_1^t, a_2^t, s_2^{0:t}~|~s_1^{0:t},a^{0:t-1})}{P(a_2^t~|~s_1^{0:t},a^{0:t-1})} \\
    & \propto \sum_{s_2^{0:t}} P(a_1^t, a_2^t, s_2^{0:t}~|~s_1^{0:t},a^{0:t-1}) \\
    &= \sum_{s_2^{0:t}} P(a_2^t~|~a_1^t, s_2^{0:t},s_1^{0:t},a^{0:t-1})\cdot P(a_1^t~|~s_2^{0:t},s_1^{0:t},a^{0:t-1}) \\ & ~~~~~~~~~~~\cdot P(s_2^{0:t}~|~s_1^{0:t},a^{0:t-1})\\
     &= \sum_{s_2^{0:t}}  \pi_2(a_2^t~|~s_2^{0:t},a^{0:t-1},a_1^t) \cdot \pi_1^*(a_1^t~|~s_1^t,s_2^t) \cdot P(s_2^{0:t}~|~s_1^{0:t},a^{0:t-1})
\end{split}
\end{equation}
Thus, we reach the policy for agent $1$:
\begin{equation}
\begin{gathered}
        \pi_1(a_1^t~|~s_1^{0:t},a^{0:t-1}, a_2^t) \propto \sum_{s_2^{0:t}}  \pi_1^*(a_1^t~|~s_1^t,s_2^t) \cdot \pi_2(a_2^t~|~s_2^{0:t},a^{0:t-1},a_1^t) \cdot  P(s_2^{0:t}~|~s_1^{0:t},a^{0:t-1})
\end{gathered}
\end{equation}
We can use the same steps to derive the symmetric policy for agent $2$:
\begin{equation}
\begin{gathered}
        \pi_2(a_2^t~|~s_2^{0:t},a^{0:t-1},a_1^t) \propto \sum_{s_1^{0:t}} \pi_2^*(a_2^t~|~s_1^t,s_2^t) \cdot \pi_1(a_1^t~|~s_1^{0:t},a^{0:t-1},a_2^t) \cdot P(s_1^{0:t}~|~s_2^{0:t},a^{0:t-1})
\end{gathered}
\end{equation}

\subsection{Extension of Theorem. \ref{theorem1} to $N$-Agent Teams}
Moving beyond two-agent teams, robots can also alternate between speaker and listener roles in $N$-agent scenarios to obtain optimal centralized performance. Let one agent be the listener, and let the other $N-1$ agents speak to this listener. As in the two-agent case, we define $a^*_i$ as the optimal centralized action for agent $i$, and we define $\bar{a}_i$ as the naive action that purely depends on agent $i$'s local state. Under our surjective mapping assumption, we have $g_i(\bar{a}_i) = s_i$. We choose the speaker $i$'s action to be $\bar{a}_i$ and the listener $j$'s action to be $a_j^* + (N-1)(a_j^* - \bar{a}_j)$. Agents alternate the listener role with interval $dt$. Taking the limit as $dt\rightarrow0$, the overall team action $a$ becomes: 
\begin{small}
\begin{eqnarray*}
\label{eq:theorem_N}
\begin{aligned}
    a^t &= \lim_{dt\rightarrow 0} \frac{\sum_{k=0}^ {N-1} a^{t+kdt}}{N} \\
        &=  {\lim_{dt \rightarrow 0}\frac{1}{N}
    \bigg(\begin{bmatrix} a_1^{*t} + (N-1)(a_1^{*t} - \bar{a}_1^t) \\ 
    \bar{a}_2^t \\ \vdots \\ \bar{a}_N^t \end{bmatrix} + \begin{bmatrix}\bar{a}_1^{t+dt}\\
     a_2^{*t+dt} + (N-1)(a_2^{*t+dt} - \bar{a}_2^{t+dt})\\ \vdots \\ \bar{a}_N^{t+dt}\end{bmatrix}} \\
     &~~~~~~+ \hdots + \begin{bmatrix}\bar{a}_1^{t+(N-1)dt}\\
     \bar{a}_2^{t+(N-1)dt} \\ \vdots \\ a_N^{*t+(N-1)dt} + (N-1)(a_N^{*t+(N-1)dt} - \bar{a}_N^{t+(N-1)dt})\end{bmatrix}\bigg)\\
     &= a^{*t}
\end{aligned}
\end{eqnarray*}
\end{small}

\noindent And so, following the simple yet effective role rotation mechanism, the decentralized team converges to the optimal action of the centralized team as role alternation frequency increases.

\subsection{Derivations for \eref{eq:L5}}
For the stochastic centralized policy, we have:
\begin{equation}
    \begin{split}
    \pi_1^*(a_1~|~s_1,s_2) = \mathcal{N}(K_{11}s_1 + K_{12}s_2, w_1^2) \\ \pi_2^*(a_2~|~s_1,s_2) = \mathcal{N}(K_{21}s_1 + K_{22}s_2, w_2^2)
\end{split}
\end{equation}

The states $s_1$ and $s_2$ are constant and independent, and sampled from Gaussian priors: $s_1\sim\mathcal{N}(\mu_1, \sigma_{s_1}^2)$ and $s_2\sim\mathcal{N}(\mu_2, \sigma_{s_2}^2)$. Without loss of generality, let agent $1$ be speaker and agent 2 is listener. The speaker and listener policies are:
\begin{equation}
    \pi_1(a_1~|~s_1) = \mathcal{N}(K_{11}s_1 + K_{12}\mu_2, \sigma_{1}^2)
\end{equation}
\begin{equation}
    \pi_2(a_2~|~s_2,a_1) = \mathcal{N}(K_{21}K_{11}^{-1}(a_1 - K_{12}\mu_2) + K_{22}s_2, \sigma_{2}^2)
\end{equation}

Now we can solve for the parameters  $\theta = \{\sigma_1, \sigma_2\}$ that minimize the KL divergence between actual and desired policies:
\begin{multline}
    \label{eq:full_KL}
    \min_{\theta} KL =  \min_{\theta}\bigg[ \log{\frac{w_1}{\sigma_1}} + \frac{\sigma_1^2 + (K_{11}s_1 + K_{12}s_2 - K_{11}s_1 - K_{12}\mu_2)^2}{2w_1^2} +  \log{\frac{w_2}{\sigma_2}} + \\ \frac{\sigma_2^2 + (K_{21}s_1 + K_{22}s_2 - K_{21}K_{11}^{-1}(a_1 - K_{12}\mu_2) - K_{22}s_2)^2}{2w_2^2}\bigg]\\
    =  \min_{\theta}\bigg[ \log{\frac{w_1w_2}{\sigma_1\sigma_2}} + \frac{\sigma_1^2}{2w_1^2} + \frac{K_{12}^{2}(s_2 - \mu_2)^2}{2w_1^2} + \frac{\sigma_2^2 + K_{21}^{2}(s_1 - K_{11}^{-1}(a_1 - K_{12}\mu_2))^2}{2w_2^2}\bigg]
\end{multline}
Taking the expectation over \eref{eq:full_KL}, we reach:
\begin{equation}
    \min_{\theta} \mathbb{E}[KL] =  \min_{\theta}\bigg[\frac{K_{12}^{2}\sigma_{s_2}^2}{2w_1^2} + \log{\frac{w_1w_2}{\sigma_1\sigma_2}} + \frac{\sigma_1^2}{2w_1^2} + \\ \frac{\sigma_2^2 +  K_{21}^{2}\sigma_1^2/K_{11}^{2}}{2w_2^2}\bigg]
\end{equation}
Thus, the optimal choices of $\sigma_1$ and $\sigma_2$ that minimize the KL divergence are:
\begin{equation}
    \sigma_1 = \frac{K_{11}w_1w_2}{\sqrt{K_{11}^2w_2^2 + K_{21}^2w_1^2}}, \quad \sigma_2 = w_2 
\end{equation}



\subsection{Simulation Environment}
In our simulations the agents move with velocity $v_i$. Formally, \eref{eq:env_dynamics} gives the system dynamics:
\begin{equation} \label{eq:env_dynamics}
        v_c^{trans} = \frac{1}{2}(v_1 + v_2), \quad  \omega = \frac{1}{r} u_i^r \times (v_i - v_c^{trans})
\end{equation}
The table's translation velocity $v_c^{trans}$ is the average of two agents' input velocities. The angular velocity is computed by first finding the difference in velocities orthogonal to the table and then dividing by $r$, the length from table center to the agents.

\subsection{Planning}
 We adopt the well-known potential field model~\cite{barraquand1992numerical} for planning. Agents at location $q$ plan their path under the influence of an artificial potential field $U(q)$, which is constructed to reflect the environment. There are two types of potential field sources. We denote the set of attractors as $k \in \mathcal{A}$ and the set of repulsive obstacles as $j \in \mathcal{R}$. The overall potential field is the sum of all attractive and repulsive potential field sources as in \eref{eq:overall_pf}. Agents move along the gradient of this potential field \eref{eq:agent_v}. $w_v$ is a predefined hyperparameter controlling scaling.
\begin{equation}
    U(q) = \sum_{k\in \mathcal{A}}U^k_{att}(q) + \sum_{j\in \mathcal{R}}U^j_{rep}(q) 
    \label{eq:overall_pf}
\end{equation}
\begin{equation}
    v_i(q_i) = -w_v\nabla U(q_i) = -w_v (\sum_{k\in \mathcal{A}_i}\nabla U^k_{att}(q_i)  +\sum_{j\in \mathcal{R}_i} \nabla U^j_{rep}(q_i))
    \label{eq:agent_v}
\end{equation}
The gradients for the attractive and repulsive fields are shown in \eref{eq:att_pf} and \eref{eq:rep_pf}. $w_{att}$ and $w_{rep}$ are hyperparameters controlling the relative scale of the attractive and repulsive fields. $q_k$ is the location for attractor $k$. $q_j$ is the location for repulsive source $j$. $\rho_j(q)$ is the minimum distance of agent at location $q$ to obstacle $j$. $\rho_0$ is a hyperparameter controlling the effective range of the repulsive potential field. The repulsive potential field is 0 outside of the range $\rho_0$.
\begin{equation}
    \nabla U^k_{att}(q) = w_{att} * \frac{q-q_k}{\left\|q-q_k\right\|}
    \label{eq:att_pf}
\end{equation}
\begin{equation}
    \nabla U^j_{rep}(q)=\left\{\begin{array}{ll}{w_{rep} * \left(\frac{1}{\rho_j(q)}-\frac{1}{\rho_{0}}\right)\left(\frac{1}{\rho_j(q)}\right) \frac{q-q_j}{\left\|q-q_j\right\|}} & {\text { if } \rho_j(q) \leq \rho_{0}} \\ {0} & {\text { if } \rho_j(q)>\rho_{0}}\end{array}\right.
    \label{eq:rep_pf}
\end{equation}

\subsection{Inference}
As discussed in \sref{sec:roles_intro} and \sref{sec:roles_use}, the listener performs inference over the speaker $i$'s action $v_i(q_i)$. Since the goal attractor is known and shared in our collaborative tasks, the listener is trying to infer the speaker's repulsive obstacles $j \in \mathcal{R}_i$. In our implementation, the listener approximates speaker $i$'s repulsive potential field with one inferred obstacle $\bar{j}_i$ at location $\bar{q}_i$. More formally, the listener solves:
\begin{equation}
    v_i(q_i) = -w_v (\sum_{k\in \mathcal{A}_i}\nabla U^k_{att}(q_i)  + \nabla U^{\bar{j}_i}_{rep}(\bar{q}_i))
    \label{eq:inference}
\end{equation}
Combined with \eref{eq:att_pf} and \eref{eq:rep_pf}, the numerical solution for obstacle $\bar{j}_i$'s location $q_{\bar{j}_i}$ is obtained by bisection iteration.

\subsubsection{Supplementary Results: Rapidly Changing Roles Outperforms Static Roles}
In addition to the success rate $\lambda$ from \tabref{table:table1}, we also measured the average game length $l$. The average game length \textit{during only failure cases} is shown in \tabref{table:homo_game_time}. Changing roles rapidly fails in games that require more time steps. This result indicates that the cases where \textit{Roles (Dynamic)} fail are the most difficult, requiring longer and more complicated trajectories to avoid the obstacles.

\begin{table}[]
\centering
\makegapedcells
\caption{Average game length in failure cases when both robots knew the geometry of the obstacles \textit{a priori}.}
\vspace{0.5em}
\begin{tabular}{c|ccccc}
      & \multicolumn{3}{c}{\textbf{Roles (Dynamic)}} & \multicolumn{2}{c}{\textbf{Roles (Static)}} \\ 
      Obstacles & \textbf{$T=1$} & \textbf{$T=4$} & \textbf{$T=16$} & Speaker-Listener & Speaker-Speaker \bigstrut    \\ \hline 
n = 2  & \textbf{164.00}     & 152.93     & 146.53     & 145.96          & 143.13          \\ 
n = 4 & \textbf{181.11}     & 154.98     & 149.99     & 148.04          & 146.75          \\ 
n = 8 & \textbf{167.61}     & 160.47     & 156.89     & 158.13          & 154.65          \\ 
\end{tabular}
\label{table:homo_game_time}
\shrink
\end{table}

\subsubsection{Supplementary Results: Roles with Noisy Action Observations Match Noisy Messages}
In \sref{sec:exp_3}, we explored noisy actions where the coefficient of variance was 0.1 (see  \tabref{table:table2}). Here we also provide the results from different coefficients of variation: $\frac{\sigma}{\mu} \in \{0.001, 0.01, 0.1\}$. Our full results are summarized in \tabref{table:full_noise_success_rate}. As noise level increases, the success rate gradually drops and the gap between dynamic roles and explicit communication becomes greater. This is because the error gets amplified by the nonlinearity in the action inference process. Nevertheless---across all cases---the dynamic role strategy dealt with noise and performed on par with explicit communication.

\vspace{1em}

\begin{table}[h!]
\shrinktop
\centering
\caption{Success rate $\lambda$ over 1000 games when communication is noisy. $\frac{\sigma}{\mu}$ is coefficient of variation that controls the noise level for both action observation and explicit communication. Here S-L represents the static Speaker-Listener team and S-S represents the static Speaker-Speaker team.}
\vspace{1em}
\makegapedcells
\begin{tabular}{c|ccccccc}
                     &    & \textbf{Explicit} & \multicolumn{3}{c}{\textbf{Roles (Dynamic)}} & \multicolumn{2}{c}{\textbf{Roles (Static)}} \\ 
Obstacles & noise level \Gape[3pt][2pt]{$\frac{\sigma}{\mu}$} &$T=0$ & $T=1$ & $T=4$ & $T=16$ & S-L & S-S \bigstrut \\  \hline 
\multirow{3}{*}{n=2} 
                     & 0.001 & 0.997 & 0.989     & 0.89      & 0.818    & 0.889    &  \multirow{3}{*}{0.758}                      \\
                     & 0.01  & 0.996 & 0.980     & 0.891     & 0.814    & 0.889    &                        \\
                     &  0.1  & 0.925 & 0.884     & 0.818     & 0.787    & 0.829    &                        \\ \hline
\multirow{3}{*}{n=4} 
                     &  0.001 & 0.984 & 0.943     & 0.778     & 0.653    & 0.731    & \multirow{3}{*}{0.558}                       \\
                     &  0.01  & 0.981 & 0.927     & 0.768     & 0.651    & 0.731    &                        \\
                     &  0.1   & 0.833 & 0.747     & 0.669     & 0.606    & 0.655    &                        \\ \hline
 \multirow{3}{*}{n=8} 
                     &  0.001 & 0.904 & 0.738     & 0.531     & 0.388    & 0.495    & \multirow{3}{*}{0.296}  \\
                     &  0.01  & 0.900 & 0.716     & 0.504     & 0.388    & 0.491    &                        \\
                     &  0.1   & 0.553 & 0.512     & 0.416     & 0.357    & 0.398    & \\ 
\end{tabular}
\label{table:full_noise_success_rate}
\shrinkbottom
\end{table}
\end{document}